\documentclass[sigconf]{acmart}
\renewcommand\footnotetextcopyrightpermission[1]{}
\usepackage{afterpage}
\usepackage{graphicx}
\usepackage{multirow}
\usepackage{multicol}
\AtBeginDocument{%
  }

\begin{document}

\title[Chasing the Timber Trail: Machine Learning to Reveal Harvest Location Misrepresentation]{Chasing the Timber Trail: \\Machine Learning to Reveal Harvest Location Misrepresentation}

\author{Shailik Sarkar}
\affiliation{%
  \institution{Computer Science, Virginia Tech}
  \city{Alexandria}
  \state{VA}
  \country{USA}
}
\email{shailik@vt.edu}

\author{Raquib Bin Yousuf}
\affiliation{%
  \institution{Computer Science, Virginia Tech}
  \city{Alexandria}
  \state{VA}
  \country{USA}
}
\email{raquib@vt.edu}

\author{Linhan Wang}
\affiliation{%
  \institution{Computer Science, Virginia Tech}
  \city{Alexandria}
  \state{VA}
  \country{USA}
}
\email{linhan@vt.edu}

\author{Brian Mayer}
\affiliation{%
  \institution{Computer Science, Virginia Tech}
  \city{Alexandria}
  \state{VA}
  \country{USA}
}
\email{bmayer@cs.vt.edu}

\author{Thomas Mortier}
\affiliation{%
  \institution{Word Forest ID}
  \city{Washington, DC}
  \country{USA}
}
\email{thomas.mortier@worldforestid.org}

\author{Victor Deklerck}
\affiliation{%
  \institution{Meise Botanic Garden}
  \country{Belgium}
}
\email{victor.deklerck@worldforestid.org}

\author{Jakub Truszkowski}
\affiliation{%
  \institution{Word Forest ID}
  \city{Washington, DC}
  \country{USA}
}
\email{jakub.truszkowski@worldforestid.org}

\author{John C. Simeone}
\affiliation{%
 \institution{Simeone Consulting, LLC}
 \city{Littleton}
 \state{New Hampshire}
 \country{USA}}
\email{simeoneconsulting@gmail.com}

\author{Marigold Norman}
\affiliation{%
  \institution{Word Forest ID}
  \city{Washington, DC}
  \country{USA}
}
\email{marigold.norman@worldforestid.org}

\author{Jade Saunders}
\affiliation{%
  \institution{Word Forest ID}
  \city{Washington, DC}
  \country{USA}
}
\email{jade.saunders@worldforestid.org}

\author{Chang-Tien Lu}
\affiliation{%
  \institution{Computer Science, Virginia Tech}
  \city{Alexandria}
  \state{VA}
  \country{USA}
}
\email{clu@vt.edu}

\author{Naren Ramakrishnan}
\affiliation{%
  \institution{Computer Science, Virginia Tech}
  \city{Alexandria}
  \state{VA}
  \country{USA}
}
\email{naren@cs.vt.edu}

\renewcommand{\shortauthors}{Sarkar et al.}

\begin{abstract}
  Illegal logging poses a significant threat to global biodiversity, climate stability, and depresses international prices for legal wood harvesting and responsible forest products trade, affecting livelihoods and communities across the globe. Stable isotope ratio analysis (SIRA) is rapidly becoming an important tool for determining the harvest location of traded, organic, products. The spatial pattern in stable isotope ratio values depends on factors such as atmospheric and environmental conditions and can thus be used for geographic origin identification. We present here the results of a deployed machine learning pipeline where we leverage both isotope values and atmospheric variables to determine timber harvest location. Additionally, the pipeline incorporates uncertainty estimation to facilitate the interpretation of harvest location determination for analysts. We present our experiments on a collection of oak (\textit{Quercus} spp.) tree samples from its global range. Our pipeline outperforms comparable state-of-the-art models determining geographic harvest origin of commercially traded wood products,
  and has been used by European enforcement agencies to identify harvest location misrepresentation. We also identify opportunities for further advancement of our framework and how it can be generalized to help identify the origin of falsely labeled organic products throughout the supply chain.
\end{abstract}

\maketitle

\section{Introduction}
Illegal logging is a global problem with serious economic, social, and ecological impacts. It is the most profitable natural resource crime with an estimated annual value between \$52 billion and \$157 billion, as well as the third largest transnational crime, after counterfeiting and drug trafficking \cite{transnational}.  Illegal logging causes destruction to forest ecosystems, the local flora and fauna, and the people whose livelihoods depend upon forests for goods and services. The effects of illegal logging are not only felt in the area where the illicit activity takes place but also impact associated trade and international markets. Two decades ago it was determined that the value of US-harvested hardwoods (e.g., maple, oak, ash, poplar) had been depressed by 7-16\% due to the amount of illegal timber in global trade \cite{seneca2004illegal}. Given the complexity of global supply chains and the environmental and societal harm that practices such as illegal resource harvesting present, it is increasingly difficult for stakeholders (e.g., consumers, businesses involved in international trade, and governments seeking to enforce regulations) to ensure that
products do not contribute to such environmental and social ills. 

Knowing whether illegal logging has taken place starts with identifying where and when the specific trees in question were harvested, and continues to other key data elements along the supply chain. The complexity of global supply chains and the lack of key data passing through the supply chain makes it near impossible to track and trace a product from consumer back to its harvest origin. Supply chain actors may try to misrepresent the harvest location of timber products for a variety of reasons, some of which may be deliberate (e.g., illegal logging or other type of illicit or fraudulent activity such as sanctions evasion), or honest mistakes in the reported geographic harvest origin (e.g., at a fine scale spatial resolution, misrepresentation due to inaccuracies in GPS signal). 
Examples of deliberate misrepresentation of harvest location include:
\begin{itemize}
    \item In 2015, US flooring importer Lumber Liquidators plead guilty to environmental crimes related to importing illegally harvested Russian wood, which included felony misdeclarations on its US import documentation, claiming the species of oak used was harvested in Germany, even though that specific species only grows in Northeast Asia \cite{office_2015}; 
    \item Due to Russia’s 2022 war in Ukraine and the sanctions regime put in place by many countries, there has been increased scrutiny on sanctions evasion of Russian timber entering Europe by misdeclaring the harvest origin \cite{earthsight_blood-stained, brussels_times}. 
\end{itemize}

What is now possible, due to global efforts including the effort described here, is to use stable isotope ratio analysis (SIRA) to validate the claimed geographic harvest origins from finished forest products such as furniture or flooring. Stable isotopes are chemical variants of elements that do not go through radioactive decay. 
By measuring stable isotope values through mass
spectrometry~\cite{barrie1996automated} and comparing
it to a location-referenced database, we can help 
understand the origin of organic products.
Previous work has shown how it can be used to trace the origin of items such as wood, seafood, agricultural products, and cotton \cite{truszkowski2025, mortier2024framework, watkinson2022case, cusa_future_2022, wang2020tracing, meier-augenstein_discrimination_2014}. 

We focus on wood and forest products, which play a critical role in the fight against climate change and, as mentioned earlier, the illegal harvesting of trees is the most profitable natural resource crime \cite{transnational, messetchkova_glasgow_2021}. The use of stable isotope testing as a tool to determine the origin of such products is seen as a critical mechanism to 
identify instances of false or misrepresented harvest location claims of wood and forest products, thus
assisting to uncover instances of illegal logging and sanctions evasion.

Our key contributions are:
\begin{enumerate}
    \item We present a comprehensive multi-task Gaussian process modeling framework that supports the incorporation of auxiliary data such as climate layers to support origin determination. This enables the
    incorporation of environmental factors, imputing
    uncertainty to predictions, and multimodal feature integration.
    \item Our work is a {\bf deployed} machine learning
    pipeline wherein physical samples are collected, subject to tests, and injected into our model to help European enforcement agencies in combating illegal timber trade by  demonstrating that a claimed harvest location other than Russia is not viable. See coverage of our work from the {\it New York Times}~\cite{nytimes_nazaryan_2024}.
    \item While (due to confidentiality reasons) we are unable to showcase specific instances of mislabeled data in this paper, we demonstrate accuracy profiles of our approach in a controlled experiment that illustrates the interplay between SIRA values and atmospheric variables and how they affect our ability to reveal harvest location misrepresentation. This goes beyond traditional ML pipelines that only predict isotope values into an end-to-end approach that supports decision-making by enforcement agencies.
\end{enumerate}

\section{Methodology}
\begin{figure*}[t]
    \centering
    \includegraphics[ scale = 0.33]{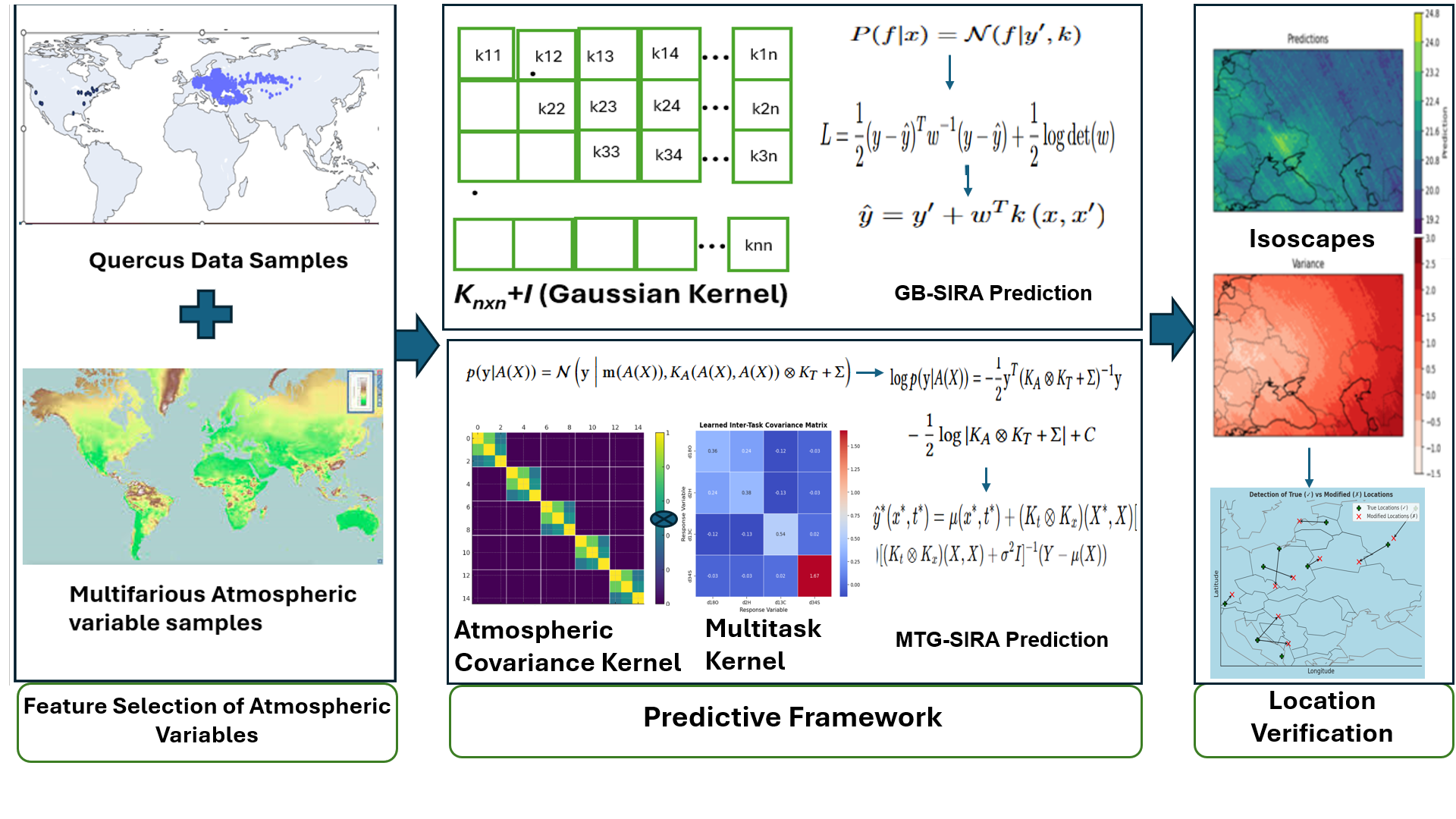}
    \vspace{-1.5\baselineskip}
    \caption{ML Pipeline for SIRA}
    \label{fig:framework}
\end{figure*}

\subsection{Stable Isotope Ratios}
A stable isotope ratio refers to the ratio of two stable isotopes of a single element~\cite{coplen1983comparison}. 
The natural variation observed for this ratio is determined by underlying mechanisms that are affected by a range of different factors including but not limited to environmental, atmospheric, soil, metabolic fraction, or other characteristics specific to a species \cite{siegwolf_stable_2022, wang2021possible, vystavna2021temperature}.
The measurement of stable isotope ratios is commonly expressed using delta (\( \delta \)) notation, which represents variations in the isotope ratio in parts per thousand (‰, per mil). The most commonly analyzed stable isotopes include those of oxygen, carbon, hydrogen, nitrogen, and sulfur, which are widely used in various scientific fields such as geology, ecology, and climate science. For each element, the delta notation indicates the deviation of the sample's isotope ratio from a standardized reference. In general, it is defined as follows:

\begin{align*}
\delta^{n} \mathrm{E} = \left( \frac{\left( \frac{^{n}\mathrm{E}}{^{n'}\mathrm{E}} \right)_{\text{sample}}}{\left( \frac{^{n}\mathrm{E}}{^{n'}\mathrm{E}} \right)_{\text{standard}}} - 1 \right) \times 1000
\end{align*}

\noindent
where \( \mathrm{E} \) represents the element of interest, \( ^{n}\mathrm{E} \) is the heavier isotope, and \( ^{n'}\mathrm{E} \) is the lighter isotope. The following are the four stable isotope ratios used in this paper:
\begin{itemize}
    \item $\delta^{13}\mathrm{C}$: a measure of the ratio of two stable isotopes of Carbon ($^{13}\mathrm{C}$ and $^{12}\mathrm{C}$). 
    \item $\delta^{18}\mathrm{O}$: a measure of the ratio of two stable isotopes of Oxygen ($^{18}\mathrm{O}$ and $^{16}\mathrm{O}$).
    \item $\delta^{2}\mathrm{H}$: a measure of the ratio of two stable isotopes of Hydrogen ($^{2}\mathrm{H}$ and $^{1}\mathrm{H}$).
    \item $\delta^{34}\mathrm{S}$: a measure of the ratio of two stable isotopes of Sulfur ($^{34}\mathrm{S}$ and $^{32}\mathrm{S}$).
\end{itemize}

In this paper, we propose an end-to-end ML Pipeline that leverages different techniques for different stages of the workflow as illustrated by Figure~\ref{fig:framework}.

\subsection{Atmospheric Data Layers}
Stable isotope ratios are largely dependent on various environmental factors, including but not limited to precipitation, water vapor pressure, and reflected shortwave radiation~\cite{vystavna2021temperature,elbeltagi2023prediction}. 
The ecological process that facilitates enrichment, or the lack thereof, for stable isotopes in oak trees is a lengthy process. This process can be influenced by climatic patterns observed over long periods. 
To capture overall climatic conditions, including regularities and anomalies, we collect comprehensive data on 25 different atmospheric properties~\cite{nasaReflectedShortwave,huffman2020gpcp,bowen2003interpolating} spanning over 20 years across the world.

\subsection{Modeling Framework} \label{subsec:prediction}
Gaussian process methods are adept at modeling spatial data but to capture complex non-linear relationships with the atmospheric variables, decision tree-based boosting algorithms have been shown to be effective. We describe a
combined Gaussian process-mixed effect
modeling approach motivated by~\cite{sigrist2022gaussian}.
Given a set of locations $X$={$x_1$,$x_2$...,$x_n$} with corresponding atmospheric features $A(X)$ = {$A(x_1)$, $A(x_1)$, ..., $A(x_n)$}, where $A(X_i)$ is the atmospheric feature at location $x_i$, we first aim to configure two separate learners, as outlined next.
We first provide a description of the mixed-effects learner, followed by an explanation of Gaussian Process Regression (GPR).

To model the non-linear relationship between atmospheric variables and stable isotope ratios, we use mixed-effect modeling comprised of multiple tree-based learners. The algorithm tries to learn a set $S$ of tree-based base learners $f(\cdot)$ in a function space $H$, defined as the linear function span of $S$. Hence, the prediction derived from this mixed-effect model is defined as $y'=F(A(X))$.

GPR is particularly popular in this context as it naturally aids in uncertainty estimation. The observed values are viewed as the realization of
a random process
defined as $ P(f|x)=\mathcal{N}(f|\mu,k) $
where $\mu =[m(x_1),m(x_2)...m(x_n)]$ is the mean estimate and $K$ is the covariance matrix given by $K_{ij}=k(x_{i},x_{j})$ where $k$ is the kernel covariance function. Traditionally, the Gaussian process is assumed to be zero mean. However, we combine the output of the base decision tree learner by modifying the probability density function in the following way:
\begin{align}
    P(f|x)=\mathcal{N}(f|y',k)
\end{align}
In this work, the Gaussian process takes an input $X=\{x_{1},x_{2},x_{3}...x_{n}\}$ where $x_{i}$ denotes the location of i\textsuperscript{th} sample in the training set. The target variable $y$ is a single isotope ratio. The GPR fits the following kernel coefficient:
\begin{align}
    w = [K_{nn}+\Sigma]^{-1}y
\end{align}
where $\Sigma$ is the error term that corresponds to the uncertainty estimation and $K_{nn}$ is the covariance matrix of dimension ($n$ x $n$) where $n$ is the number of samples in the training set. 

Now, for a new set of unseen samples $x$, the prediction is calculated as:
\begin{align}
    \hat{y} = y'+w^{T}k\left(x, x^{\prime}\right)
\end{align}
In our case, $K$ is the kernel function of our choosing. We use a combination of radial basis function, periodic, and rational quadratic function as the covariance function as shown below:
\begin{align}
k\left(x, x^{\prime}\right)=\begin{bmatrix}
\exp \left(-\frac{\left(x-x_{1}\right)^2}{2 \ell^2}\right)+\left(1+\frac{\left(x-x_{1}\right)^2}{2 \alpha \ell^2}\right)^{-\alpha}\\+ \exp \left(-\frac{2 \sin ^2\left(\pi\left|x-x_{1}\right| / p\right)}{\ell^2}\right),\\
\exp \left(-\frac{\left(x-x_{2}\right)^2}{2 \ell^2}\right)+\left(1+\frac{\left(x-x_{2}\right)^2}{2 \alpha \ell^2}\right)^{-\alpha}\\+ \exp \left(-\frac{2 \sin ^2\left(\pi\left|x-x_{2}\right| / p\right)}{\ell^2}\right),\\
.\\
.\\
.\\
\exp \left(-\frac{\left(x-x_{n}\right)^2}{2 \ell^2}\right)+\left(1+\frac{\left(x-x_{n}\right)^2}{2 \alpha \ell^2}\right)^{-\alpha}\\+ \exp \left(-\frac{2 \sin ^2\left(\pi\left|x-x_{n}\right| / p\right)}{\ell^2}\right)
\end{bmatrix}
\end{align}

\subsection{Joint Optimization}
As described in \cite{sigrist2022gaussian} we aim to combine the two learners through a joint optimization goal. Here the loss function is the negative log-likelihood function as described below:
\begin{align}
    L=\frac{1}{2}(y-\hat{y})^T w^{-1}(y-\hat{y})+\frac{1}{2} \log \operatorname{det}(w)
\end{align}
If $\mathbf{w}$ is the parameter of the Gaussian Process co-efficient kernel and $F(\cdot)$ is the mixed-effect learner  $F(A(X))$ the goal is to find the joint minimizer:
\begin{align}
    (\hat{F}(\cdot), \hat{\mathbf{w}})=\underset{(F(\cdot), \mathbf{w}) \in(\mathcal{H}, \mathbf{w})}{\operatorname{argmin}} R(F(\cdot), \mathbf{w})
\end{align}
where $R(F(\cdot), \mathbf{w})$ is a risk functional defined as:
\begin{align}
    R(F(\cdot), \mathbf{w}):\quad(F(\cdot), \mathbf{w}) \mapsto L
\end{align}
Here, $L$ is the loss function described earlier. R is determined by evaluating $F(A(X))$, $\hat{y}$ and then calculating the loss function $L$. The risk factor $R$ is minimized using the Gaussian Process boosting algorithm. 
First, for iteration $i$ we determine $\mathbf{w}_{i}$ as the following:
\begin{align}
    \mathbf{w}_{i}=\operatorname{argmin}_{\mathbf{w} \in \mathbf{w}} L\left(y, F_{i-1}, \mathbf{w}\right)
\end{align}
where $L$ is the loss function denoted in equation 5.
Then, we update base learner $F_{i}$ through a  Newton step:
\begin{align}
   F_{i}= F_{i-1}-\frac{R^{\prime}\left(F_{i-1}\right)}{R^{\prime \prime}\left(F_{i-1}\right)}
\end{align}
where $R^{\prime}$ and $R^{\prime \prime}$ denote first and second-order derivatives of the reward function for the $(i-1)$\textsuperscript{th} iteration of F.

\subsection{Modeling Dependencies between Isotopes}
Although multivariate Gaussian Processes in the form of co-kriging have been explored for stable isotope prediction, the idea of inter-task dependencies has not been explored yet and is novel to our approach. Due to the strong correlations that stable isotope ratios often exhibit, we hypothesize that using a Gaussian process to leverage both the spatial correlation of atmospheric variables and the inter-task dependencies of different stable isotopes could greatly benefit SIRA prediction. Further, we aim to leverage automatic relevance judgments from amongst the multivariate input of atmospheric variables.
Similar to the GPR model, the Gaussian Process for the multi-task model defines a distribution over the function f defined over the space spanned by atmospheric vector A(x):

\begin{equation}
\begin{aligned}
   \mathbf{f}(A(x)) &= 
   \begin{bmatrix} 
      f_{T_1}(A(x)), \dots, f_{T_M}(A(x)) 
   \end{bmatrix}^T \\
   &\sim \mathcal{GP}\big(m_m(A(x)), k_m(x, x')\big)
\end{aligned}
\end{equation}

\noindent
where $f_{T_i}(A(x))$ can be viewed as the objective function for task i, with the task corresponding to one of the four stable isotope ratio being modeled. \( m_m(A(x)) \) is the mean function for the atmospheric feature at location x and \( k_m(x,x') \) is the multitask covariance function for location x and x'. The covariance between the feature vector A(X) across multiple tasks is given by a Kronecker product structure:

\begin{equation}
    k_{MT}( (A(x), T_{i}), (A(x'), T_{j}) ) = K(A(x), A(x')) \otimes K_{T_{i}, T_{j}}
\end{equation}

\noindent
where \( K(A(x), A(x')) \) represents the covariance function for the atmospheric vector corresponding to location x and x' and \( K_{T_{i}, T_{j}} \) is the task covariance matrix between two tasks $T_{i}$ and $T_{j}$.  

\subsection{Interpretability via the Covariance Matrix}
The feature covariance matrix, i.e., K(A(x), A(x')), is parameterized by automatic relevance determination of the length scale for the chosen covariance matrix. As Matern kernels have been proven to be effective in leveraging the sudden change in the latent feature space, and an RBF kernel is effective in capturing a smoother correlation in the feature space, we use
a combination of these two kernels to constitute our covariance matrix:

\begin{align}
K(A(x), A(x')) &= \lambda_1 K_{\text{RBF}}(A(x), A(x')) + \lambda_2 K_{\text{Matérn}}(A(x), A(x'))
\end{align}
\begin{align}
K_{\text{RBF}}(A(x), A(x')) &= \sigma^2 \exp\left( -\frac{1}{2} \sum_{d=1}^{D} \frac{(A_d(x) - A_d(x'))^2}{\ell_{T_i, d}^2} \right) \\[4pt]
K_{\text{Matérn}}(A(x), A(x')) &= \sigma^2 \prod_{d=1}^{D} 
    \left( 1 + \frac{\sqrt{3} | A_d(x) - A_d(x') |}{\ell_{T_i, d}} \right) \notag \\[8pt]
    &\quad  \exp\left( -\frac{\sqrt{3} | A_d(x) - A_d(x') |}{\ell_{T_i, d}} \right)
\end{align}
In the above,
$\sigma^2$ is the variance,
$\ell_{T_i, d}$ is the ARD lengthscale for task $i$ and atmospheric variable at dimension $d$,
and $|A(x) - A(x')|$ denotes the Euclidean distance between the feature vector at location $x$ and $x'$.
Through optimizing the objective function we learn the ARD lengthscale value for each dimension of the feature space which is used to extract feature importance:
\begin{align}
    \mathrm{Feature\,\, Importance\,\,}{T_{i}}(d) &=\frac{1}{\ell_{T_i, d}}
\end{align}\label{eq:feature}

\noindent
In addition to feature-based interpretability, we also leverage task-specific insights through factorization of the task covariance matrix $K_{T}$:
\begin{align}
     \mathbf{K}_{T} &= \mathbf{L} \mathbf{L}^{T}
\end{align}
\noindent
where L is a lower triangular matrix.

\subsection{Multitask Loss Function}
Unlike traditional GPR we formulate a joint likelihood function over all the response variables defined as:
\begin{align}
p(\mathbf{Y} | A(X)) &= \mathcal{N} \left( \mathbf{Y} \; \Big| \; \mathbf{m}(A(X)), K_A(A(X), A(X)) \otimes K_T + \mathbf{\Sigma} \right) 
\end{align}
and the marginal log-likelihood derived as:
\begin{align}
\log p(\mathbf{Y} | A(X)) &= -\frac{1}{2} \mathbf{Y}^T (K_A \otimes K_T + \mathbf{\Sigma})^{-1} \mathbf{Y} \notag \\[8pt]
& \quad - \frac{1}{2} \log |K_A \otimes K_T + \mathbf{\Sigma}| + C
\end{align}
where C is a normalization constant.
\subsection{Location Verification}
Given $Y_{true}$, the given value of the isotope ratio, our objective is to determine whether the claim about the origin being $x_{mod}$ is true or not. 
First, we compute the likelihood of observing $Y_{true}$ for  $x_{mod}$ :
\begin{align}
    \log p(Y_{\text{true}} | A_{\text{mod}}) &= -\frac{1}{2} (Y_{\text{true}} - \mathbf{\mu}_{\text{mod}})^T \mathbf{\Sigma}_{\text{mod}}^{-1} (Y_{\text{true}} - \mathbf{\mu}_{\text{mod}}) \notag \\[5pt]
& \quad - \frac{1}{2} \log |\mathbf{\Sigma}_{\text{mod}}| - \frac{M}{2} \log 2\pi
\end{align}
Here, M is the number of isotope ratios or length of the $Y_{true}$ vector which is also the degrees of freedom for the subsequent hypothesis test; ${\mu}_{\text{mod}}$ is the predicted mean for the location $x_{mod}$ given atmospheric variable $A_{mod}$; and ${\Sigma}_{\text{mod}}$ is the corresponding covariance matrix.

We can now determine the veracity of the claim by the following hypothesis test:
\begin{itemize}
    \item $H_{0}$: The test sample $Y_{true}$ comes from the location $x_{mod}$
    \item $H_{1}$: The test sample $Y_{true}$ does not come from the location $x_{mod}$ 
\end{itemize}
Leveraging the trained GP model we formulate:
\begin{align}
    \chi^2 &= (Y_{\text{true}} - \mathbf{\mu}_{\text{mod}})^T \mathbf{\Sigma}_{\text{mod}}^{-1} (Y_{\text{true}} - \mathbf{\mu}_{\text{mod}}) 
\end{align}
According to the null hypothesis, $\chi^2$ follows a chi-squared distribution with M degrees of freedom which is the number of stable isotope ratios being modeled. The p-value can be expressed as :
\begin{align}
    p_{\text{mod}} &= P(\chi^2 > \chi^2_{\text{obs}})
\end{align}
Finally, the decision rule for classifying a claim as false is upper bound by a significance threshold $\alpha$:
\begin{align}
    \text{Decision:} & \quad p_{\text{mod}} < \alpha \Rightarrow \text{Location Claim False}
\end{align}

\section{Experimental Results}

\subsection{Datasets}
As described earlier, while our system is deployed, we are unable to present results due to confidentiality constraints. We describe results on a different dataset gathered from samples of the genus
\textit{Quercus} from countries such as China, the United States, Ukraine, and Russia. Stable isotope ratio measurements for each sample were calculated and aggregated as described in~\citep{watkinson2020development}. Each entry contained stable isotope ratio measurements of oxygen, hydrogen, sulfur, and carbon (along with their GPS coordinates).
Our atmospheric data includes isotopic composition of precipitation, water vapour, shortwave radiation, temperature, and many more factors~\cite{bowen2003interpolating,huffman2020gpcp,nasaReflectedShortwave}.
In overall, we use 25 atmospheric variables for 20 years. In mapping the atmospheric variables to the sampled location of the \textit{Quercus} spp. values, we found six of the atmospheric variables to have more than 50\% missing values due to the sparse nature of the data; these variables are discarded. This retains 19 atmospheric variables, each having 12 months of data for every year, for 20 years.

\subsection{Research Questions}
Our goals in this study are to characterize the
effectiveness of our methodology for
modeling stable isotope values and subsequent location claim verification:
\begin{itemize}
    \item \textbf{RQ1:} How does the proposed framework compare to existing approaches to stable isotope ratio prediction?
    \item \textbf{RQ2:} How do different architectural choices (such as, combining tree boosting with Gaussian processes, multivariate GP modeling, or 
    modeling inter-task correlation) benefit the outcome compared with using any of these techniques individually?
    \item \textbf{RQ3:} Does the incorporation of atmospheric variables and the feature selection methodology contribute to
    more accurate forecasts?
    \item \textbf{RQ4:} What implicit explainability can the architecture provide for atmospheric variables and inter-task correlation?
    \item \textbf{RQ5:} How effective is the proposed framework at successfully trapping false claims of timber origin?
\end{itemize}

\subsection{RQ1: Baseline Experiments}
We compare our methodology (MTG-SIRA to denote
the full multi-task framework; and GB-SIRA to denote our approach minus the multi-task formulation)
against the following alternative approaches:
\begin{itemize}
    \item \citet{watkinson2020development}: This work proposes the use of ordinary kriging for spatial interpolation of the isotope ratio values. They were used to produce predictions for carbon, oxygen, hydrogen, and sulfur stable isotope ratios on \textit{Quercus} spp., similar to our problem statement but only on samples from the United States. However, we use their methodology on our data for comparison.
    \item \citet{watkinson2022case}: This work uses Universal Kriging to predict the isotope ratio values of timber in Solomon Islands. 
    \item \citet{truszkowski2025}: This work, similar to ours, uses atmospheric data for stable isotope ratio prediction but the covariance matrix is modified by using location-specific atmospheric variables making it analogous to a co-kriging method.
    \item Support vector regression (SVR): This is a regression model variation of SVM that has shown to be effective extensively for a dataset of this size. This approach also serves as a traditional ML baseline.
    \end{itemize}
\begin{table*}[h]
 \caption{Baseline comparison of predictive models based on $R^{2}$ and RMSE for SIRA values.}
    \label{tab:baseline}
    \centering
    \setlength{\tabcolsep}{6pt}  
    \renewcommand{\arraystretch}{1.2}  
    \begin{tabular}{l c cccc cccc}
    \toprule
    \multirow{2}{*}{Baseline Comparison} & \multirow{2}{*}{Atmospheric Variables} & \multicolumn{4}{c}{$R^{2}$} & \multicolumn{4}{c}{RMSE} \\
    \cmidrule(lr){3-6} \cmidrule(lr){7-10}
    & & $\delta^{18}\mathrm{O}$ & $\delta^{13}\mathrm{C}$ & $\delta^{2}\mathrm{H}$ & $\delta^{34}\mathrm{S}$ 
    & $\delta^{18}\mathrm{O}$ & $\delta^{13}\mathrm{C}$ & $\delta^{2}\mathrm{H}$ & $\delta^{34}\mathrm{S}$ \\
    \midrule
    \citet{watkinson2020development} & -- & 0.470 & 0.320 & 0.700 & \textbf{0.690} & -- & -- & -- & -- \\
    \citet{watkinson2022case} & -- & 0.856 & 0.301 & 0.730 & 0.601 & 0.673 & 0.779 & 6.112 & 1.233 \\
    \citet{truszkowski2025} & X & 0.869 & \textbf{0.331} & 0.790 & 0.667 & 0.631 & 0.757 & 6.279 & 1.070 \\
    SVR & X & 0.754 & 0.313 & 0.750 & 0.682 & 0.940 & 0.773 & 6.410 & 1.040 \\
    GB-SIRA & X & 0.878 & 0.322 & 0.840 & 0.689 & \textbf{0.742} & \textbf{0.768} & \textbf{5.911} & \textbf{1.011} \\
    MTG-SIRA & X &\textbf{0.899} & 0.310 & \textbf{0.858} & 0.673 & \textbf{0.687} & \textbf{0.679} & \textbf{5.201} & 1.029 \\
    \bottomrule
    \end{tabular}
   
\end{table*}

\begin{table*}[h]
\caption{Ablation Study for various model choices evaluated using $R^{2}$ and RMSE.}
\label{tab:Ablation}
    \centering
    \setlength{\tabcolsep}{8pt}  
    \renewcommand{\arraystretch}{1.3}  
    \begin{tabular}{l cccc}
    \toprule
    \textbf{Ablation Study} & \multicolumn{4}{c}{\textbf{$R^{2}$}} \\
    \cmidrule(lr){2-5}
    & $\delta^{18}\mathrm{O}$ & $\delta^{13}\mathrm{C}$ & $\delta^{2}\mathrm{H}$ & $\delta^{34}\mathrm{S}$  \\
    \midrule
    \textbf{Boosting} & 0.865 & 0.329 & 0.819 & 0.675 \\
    \textbf{GPR} & 0.856 & 0.301 & 0.730 & 0.601 \\
    \textbf{Boosting + GPR} & 0.878 & 0.322 & 0.840 & 0.689  \\
    \textbf{Multivariate GP} & 0.880 & 0.299 & 0.834 & 0.625 \\
    \textbf{Multivariate and Multitasking GP} & \textbf{0.899} & 0.310 & \textbf{0.858} & 0.673  \\
    \midrule
    \textbf{Ablation Study} & \multicolumn{4}{c}{RMSE} \\
    \midrule
    \textbf{Boosting} & 0.801 & 0.780 & 5.820 & 1.070 \\
    \textbf{GPR} & 0.773 & 0.779 & 6.112 & 1.233 \\
    \textbf{Boosting + GPR} & \textbf{0.742} & \textbf{0.768} & \textbf{5.911} & \textbf{1.011} \\
    \textbf{Multivariate GP} &  0.691& 0.725 & 5.890 & 1.033 \\
    \textbf{Multivariate and Multitasking GP} &\textbf{0.687} & \textbf{0.679} & \textbf{5.201} & 1.029  \\
    \bottomrule
    \end{tabular}

\end{table*}

The experiments as detailed in Table~\ref{tab:baseline} demonstrate that, in
terms of $R^2$, our proposed frameworks GB-SIRA and MTG-SIRA 
outperform other approaches for 2 out of the
4 isotope ratio prediction tasks.
However, even for $\delta^{13}\mathrm{C}$ 
and $\delta^{34}\mathrm{S}$,
the difference with respect to the best-performing model is negligible. For isotopes such as $\delta^{2}\mathrm{H}$ and $\delta^{18}\mathrm{O}$, there is significant improvement witnessed.

In terms of RMSE, we also observe the lowest RMSE value for MTG-SIRA for all stable isotope ratios except $\delta^{34}\mathrm{S}$ for which GB-SIRA performs the best. 
We note that \citet{watkinson2020development} did not report RMSE.
The performance of both GB-SIRA and MTG-SIRA overall compares favorably, demonstrating
the effectiveness of the proposed ML framework.
In general, the performance for $\delta^{34}\mathrm{S}$ and $\delta^{13}\mathrm{C}$ is not high
across the models, a result in line with past related work. Finally, the overall performance improvement on the multitask model MTG-SIRA indicates the effectiveness of modeling inter-task dependency.

\subsection{RQ2: Architectural Choices}
To answer this research question we ablate each individual component after the feature selection module to compare against the final model. The result of this experimental setting is described in Table~\ref{tab:Ablation}. The framework without the Gaussian Process element becomes a decision-tree boosting algorithm called lightGBM which takes as input the output of the feature selection module and doesn't model the spatial element. On the other hand, without the boosting algorithm combined, GB-SIRA only takes the sample longitude and latitude as input which makes it the traditional GPR. Multivariate GP is the variation of the model that models spatial correlation by using the atmospheric variable as latent feature space. 
For the boosting algorithm, we observe encouraging performance on $\delta^{34}S$ and $\delta^{13}C$. However, when combined with Gaussian processes, the performance invariably increases. Similarly, even though multivariate Gaussian process modeling shows encouraging performance on $\delta^{18}O$, the approach of incorporating a multitask-learning module does improve the overall performance of the model. Hence, given the importance of task-specific interpretability and uncertainty estimation, MTG-SIRA proves to be the best iteration of our proposed framework. However, this is a decision to be made in consultation with all stakeholders.
\subsection{RQ3: Role of Atmospheric Variables}
To address this question, we look at the two best-performing models from Table~\ref{tab:baseline}. Since both these models take advantage of atmospheric variables, we create an experimental setting that facilitates both the inclusion and exclusion of the feature selection module. The result as described in Table~\ref{tab:feature} shows that the inclusion of the feature selection step before modeling produces a better prediction for both models. Furthermore, we also show the experimental results with and without the atmospheric variables in Table~\ref{tab:baseline}. There is a noticeable improvement for the prediction of oxygen and hydrogen stable isotope ratios.

\begin{table*}[h]
\caption{Comparison of MTG-SIRA and GB-SIRA with and without Feature Selection}\label{tab:feature}
    \centering
    \setlength{\tabcolsep}{8pt}  
    \renewcommand{\arraystretch}{1.3}  
    \begin{tabular}{l cccc cccc}
    \toprule
    \multirow{2}{*}{\textbf{Method}} & \multicolumn{4}{c}{\textbf{$R^{2}$}} & \multicolumn{4}{c}{\textbf{RMSE}} \\
    \cmidrule(lr){2-5} \cmidrule(lr){6-9}
    & $\delta^{18}\mathrm{O}$ & $\delta^{13}\mathrm{C}$ & $\delta^{2}\mathrm{H}$ & $\delta^{34}\mathrm{S}$ 
    & $\delta^{18}\mathrm{O}$ & $\delta^{13}\mathrm{C}$ & $\delta^{2}\mathrm{H}$ & $\delta^{34}\mathrm{S}$ \\
    \midrule
    \textbf{MTG-SIRA without Feature Selection} & 0.860 & 0.271 & 0.816 & 0.632 & 0.683 & 0.801 & 6.350 & 1.210 \\
    \textbf{GB-SIRA without Feature Selection} & 0.874 & 0.289 & 0.781 & 0.655 & 0.662 & 0.813 & 6.210 & 1.194 \\
    \textbf{MTG-SIRA with Feature Selection} & \textbf{0.899} & 0.310 & \textbf{0.858} & 0.673 & \textbf{0.687} & \textbf{0.679} & \textbf{5.201} & 1.029 \\
    \textbf{GB-SIRA with Feature Selection} &  0.878 & 0.322 & 0.840 & 0.689 & \textbf{0.742} & \textbf{0.768} & \textbf{5.911} & \textbf{1.011} \\
    \bottomrule
    \end{tabular}

\end{table*}

 \subsection{RQ4: Explainability}
\begin{figure*}[h]
    \centering
    \includegraphics[width=0.95\linewidth]{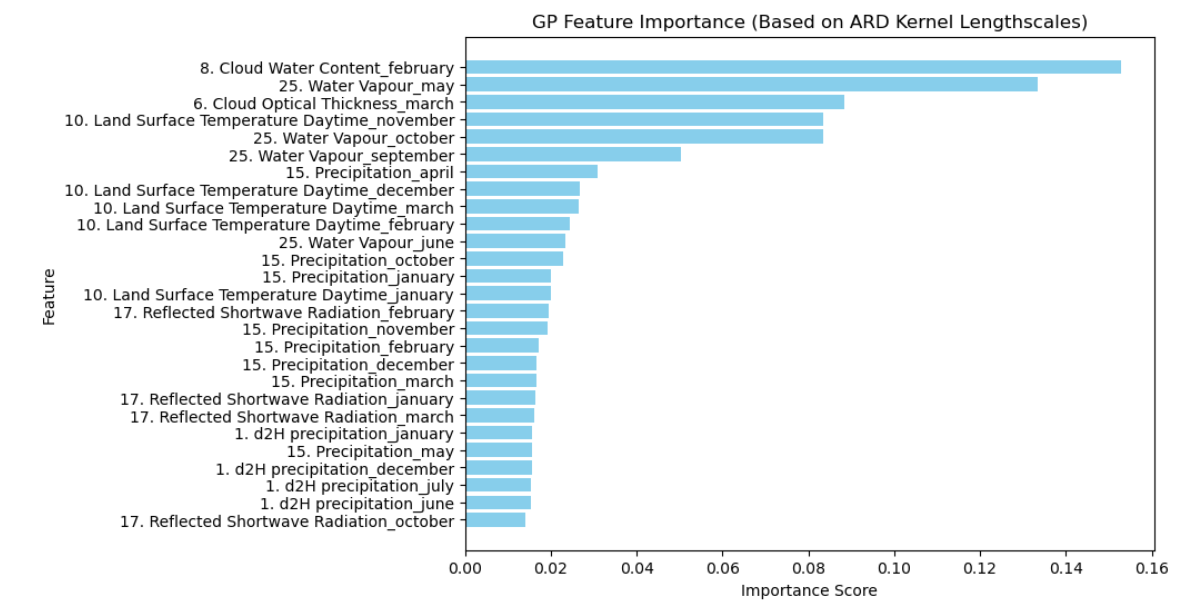}  
    \vspace{-1\baselineskip}
    \caption{Atmospheric Feature Importance}
    \label{fig:feat_imp}
\end{figure*}

\begin{figure}[h]
    \centering
    \includegraphics[width=0.95\linewidth]{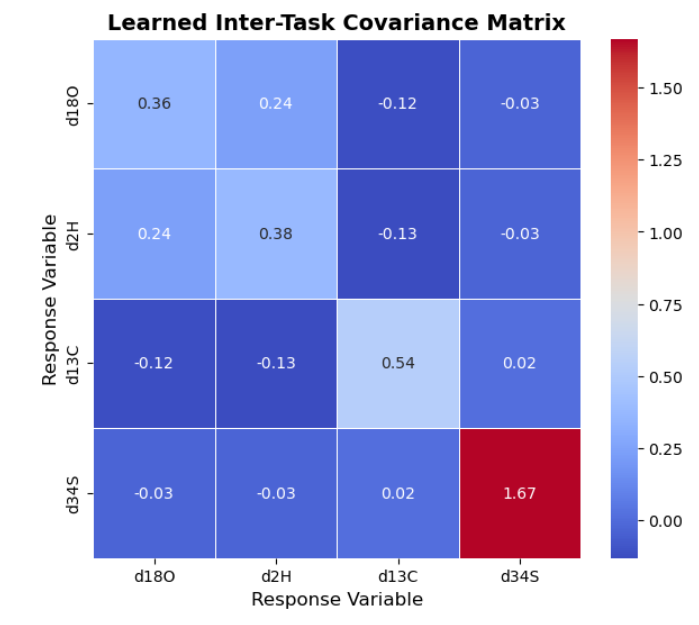}  
    \caption{Inter-task Dependency}
    \label{fig:task_imp}
\end{figure}

 As discussed in Equations~\ref{subsec:prediction} and~\ref{eq:feature}, feature importance is inversely proportional to learned lengthscales corresponding to each variable. Therefore, the trained model for MTG-SIRA can be used to visualize the feature importance of atmospheric variables for stable isotope prediction. As shown in Fig~\ref{fig:feat_imp}, the model is able to identify variables such as ``cloud water content'', ``water vapour'', ``land surface temperature'' and ``precipitation'' as some of the most important features. This is consistent with existing literature~\cite{elbeltagi2023prediction} on the subject that finds strong correlation between certain stable isotope values and the aforementioned atmospheric variables. The model's ability to identify such factors reiterates the effectiveness of MTG-SIRA as a self-explainable mechanism for feature-level interpretation. 
 Similarly, we aim to factorize the inter-task dependency and use the lower triangular matrix to gain insights into how different elemental stable isotope ratios are correlated with each other. Figure~\ref{fig:task_imp} shows strong indication of a positive correlation between $\delta^{18}O$ and $\delta^{2}H$ and a negative correlation between $\delta^{34}S$ and other stable isotopes. This can inform future usage of stable isotope measurements for location determination.

\subsection{RQ5: Evaluating Claims of Timber Origin}
As mentioned earlier, our approach is in use by European enforcement agencies. However, 
because we cannot showcase results on real data, we formulate a controlled experiment wherein we modify
the true (ground truth) location of a sample by a specified distance $d$. Thus $x$ is modified, resulting in a modified atmospheric vector for $x$. This can be formulated as: 
\begin{align}
x_{\text{mod}} &= x + \Delta x, \quad \Delta x \sim \mathcal{N}(0, \sigma_{\text{perturb}}^2) = d \\[10pt]
A_{\text{mod}} &= A(x_{\text{mod}})
\end{align}
Given $Y_{true}$ as the true value of the isotope ratio, our objective is to determine whether the claim about the origin being $x_{mod}$ is true or not. 
Then we calculate the accuracy in identifying the false claims for each value of d ranging from 500 to 5000 km in geodesic distance.

\begin{figure}[h]
    \centering
    \includegraphics[width=0.95\linewidth]{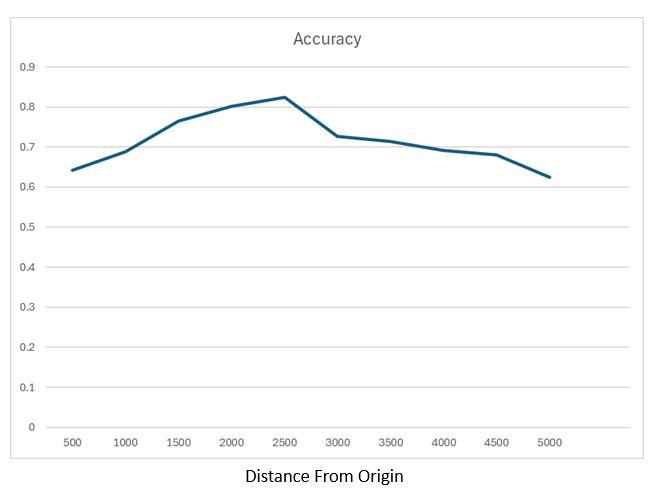}  
    \caption{Success Rate of identifying false claim vs distance from true origin}
    \label{fig:adversarial}
\end{figure}

Fig~\ref{fig:adversarial} showcases the results.
When the altered distance is within
500km, the success rate is not high (approximately 60\%),
consistent with the fact that atmospheric variables
likely do not shift significantly unless elevation
is accounted for. From this point, we begin to see
a gradual improvement in false claim detection as the distance d increases up until 2500 km, reaching an accuracy of over 80\%. As the density of training samples is high in specific areas, when the modified location falls outside the range of low uncertainty estimates for the model, we see a steady decline in the accuracy of a verification claim, ultimately becoming stable after 4000 km. This is a significant finding that can guide authorities to the responsible use of our approach for claim verification.

\section{Description of Deployment}

\begin{figure}[h]
    \centering
    \includegraphics[scale=  0.5]{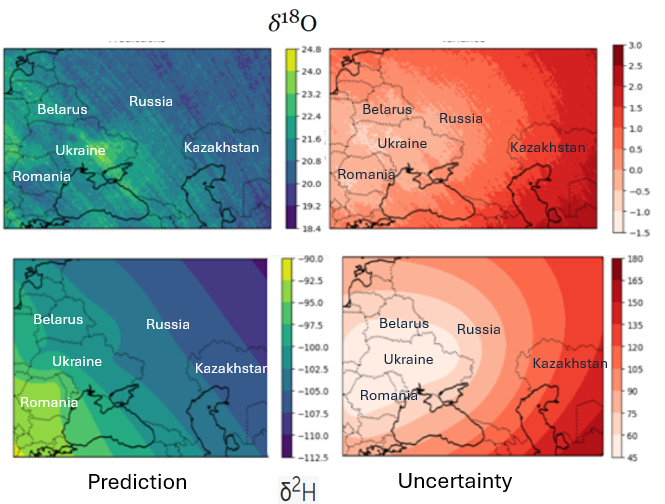}
    \vspace{-1.75\baselineskip}
    \caption{Case Study: Isoscapes for Eastern European Region-Ukraine/Russia Border for timber}
    \label{fig:case-study}
\end{figure}

Our framework is in use by European enforcement agencies to assist in demonstrating that a claimed harvest location other than Russia is not viable. See coverage of our work from the \textit{New York Times}~\cite{nytimes_nazaryan_2024}.
Due to confidentiality reasons, we present
results here on a different 
species 
(Oak; \textit{Quercus} spp.). For instance,
see 
isoscape predictions around the Russia/Ukraine border region in
Fig.~\ref{fig:case-study}.

\section{Conclusion}
We have presented a novel ML framework that uses SIRA 
to determine the origin of organic products.
It serves as a holistic end-to-end predictive pipeline that can facilitate feature selection and systematic analysis of recorded and predicted isotope ratio values. 
Our approach to multimodal feature integration
also enables us to leverage the vast information present in atmospheric data by systematic feature selection methods. Further, such incorporation of atmospheric variables into the methodology supports interpretability that prior methods do not support.

A key real-world implication of our methodology is that it can create species-wide isoscapes that predict sub-national variability even in areas without ground truth sample data. This is particularly useful if the security context makes sample data collection challenging or impossible.
Furthermore, the actionability of these species' isoscapes can be improved by visualizing uncertainty estimations to communicate the relative confidence of a predicted value in a particular region. This enables real-world decision-making based on predictions with low uncertainty values to manage legal and financial liabilities arising from timber supply contracts and law enforcement activities. 

Using our methodology, being able to infer the origins of an unseen sample with varying degrees of confidence has proven to be an effective tool used to enforce timber trade regulations and sanctions, such as the sanction on importing Russian and Belarusian timber in the EU \cite{mortier2024framework, truszkowski2025}.

There is promise for additional social and environmental impact of this work, given that the application of our SIRA methods can be applied to origin testing for a variety of organic products. Our work is particularly relevant given the
EU's Regulation on Deforestation-free Products (EUDR)
which requires proving that products made from forest-risk commodities, e.g., cattle, wood, cocoa, soy, coffee, palm oil, and rubber, do not originate from recently deforested land
or have contributed to forest degradation \cite{european_commission_regulation_2023}. 

Furthermore, there has been increased scrutiny on the use of forced labor to produce products and food that enter the global supply chain. Products such as garments and apparel made from cotton produced in the Xinjiang region of China by Uyghurs, and fish and seafood harvested by forced labor, have particularly complex supply chains \cite{mclymore_banned_2024, masters_us_2023, pbs_news_investigation_2023, cusa_future_2022}. This work contributes to the advancement of accurate scientific methods used to assist the determination of origin claims for all organic products covered in policies such as the EUDR and the US Lacey Act and Uyghur Forced Labor Prevention Act.

\bibliographystyle{ACM-Reference-Format}
\bibliography{bibliography}

\end{document}